\def\year{2019}\relax
\documentclass[letterpaper]{article} 
\usepackage{aaai19}  
\usepackage{times}  
\usepackage{helvet}  
\usepackage{courier}  
\usepackage{url}  
\usepackage{graphicx}  
\usepackage{amsmath}
\usepackage{amsthm}
\usepackage{amsfonts}
\usepackage{algorithm,algorithmic}
\usepackage{multirow}
\usepackage{subcaption}

\begin{document}
%
\title{Variational Collaborative Learning for User Probabilistic Representation}
\author{
Kenan Cui,
Xu Chen,
Jiangchao Yao, 
Ya Zhang,
\\ 
Cooperative Madianet Innovation Center\\
Shanghai Jiao Tong University\\
\{conancui, xuchen2016@sjtu.edu.cn, sunarker, ya\_zhang\}@sjtu.edu.cn}
\maketitle
\begin{abstract}

Collaborative filtering~(CF) has been successfully employed by many modern recommender systems. Conventional CF-based methods use the user-item interaction data as the sole information source to recommend items to users. However,  CF-based methods are known for suffering from cold start problems and data sparsity problems. Hybrid models that utilize auxiliary information on top of interaction data have increasingly gained attention. A few "collaborative learning"-based models, which tightly bridges two heterogeneous learners through mutual regularization, are recently proposed for the hybrid recommendation. 
However, the "collaboration" in the existing methods are actually asynchronous due to the alternative optimization of the two learners. Leveraging the recent advances in variational autoencoder~(VAE), we here propose a model consisting of two streams of mutual linked VAEs, named variational collaborative model~(VCM). Unlike the mutual regularization used in previous works where two learners are optimized asynchronously, VCM enables a synchronous collaborative learning mechanism. Besides, the two stream VAEs setup allows VCM to fully leverages the Bayesian probabilistic representations in collaborative learning. Extensive experiments on three real-life datasets have shown that VCM outperforms several state-of-art methods.

\end{abstract}

\section{Introduction}

With the rapid growth of information online, recommender systems have been playing an increasingly important role in alleviating the information overload. Existing models for recommender systems can be broadly classified into three categories~\cite{adomavicius2005toward}: content-based models, CF-based models, and hybrid models. The content-based models~\cite{lang1995newsweeder,pazzani1997learning} recommend items similar to what the user liked in the past utilizing user profiles or item descriptions. CF-based methods~\cite{Salakhutdinov:2007:PMF:2981562.2981720,He:2017:NCF:3038912.3052569,Liang:2018:VAC:3178876.3186150} model user preferences based on historic user-item interactions and recommend what people with similar preference have liked. Although CF-based models generally achieve higher recommendation accuracy than content-based methods, their accuracy drops significantly in the case of sparse interaction data. Therefore, hybrid methods~\cite{li2011generalized,Wang:2011:CTR:2020408.2020480}, utilizing both interaction data and auxiliary information, have been largely adopted in real-world recommender systems.

Collaborative Deep Learning~(CDL)~\cite{Wang:2015:CDL:2783258.2783273} and Collaborative Variational Autoencoder~(CVAE)~\cite{Li:2017:CVA:3097983.3098077} have recently been proposed as unified models to integrate interaction data and auxiliary information and shown promising results. Both methods leverage Probabilistic matrix factorization~(PMF)~\cite{Salakhutdinov:2007:PMF:2981562.2981720} to learn user/item latent factors from interaction data through point estimation. At the meanwhile, a stacked denoising autoencoder~(SDAE)~\cite{vincent2010stacked} (or a VAE~\cite{kingma2013auto}) is employed to learn latent representation from the auxiliary information. The two learners are integrated through mutual regularization, i.e., the latent representation in SDAE/VAE and the corresponding latent factor in PMF are used to regularize with each other.
However, the two learners are actually optimized alternatively, making the "collaboration" asynchronous: one-directional regularization in any iteration. Besides, due to the point estimation nature of latent factors in PMF, the regularization here fails to fully leverage the Bayesian representation of the latent variable from SDAE/VAE.

To address aforementioned problems, we propose a deep generative probabilistic model under the collaborative learning framework named \textbf{v}ariational \textbf{c}ollaborative \textbf{m}odel for user preference~(VCM). The overall architecture of the model is illustrated in Figure~\ref{fig::framework}. Two parallel extended VAEs are collaboratively employed to simultaneously learn comprehensive representations of user latent variable from user interaction data and auxiliary review text data.

Unlike CVAE and CDL, which learn separate user/item latent factors with point estimation nature through PMF, the VCM use VAE for CF~\cite{Liang:2018:VAC:3178876.3186150} to efficiently infer the variational distribution from interaction data as the \textbf{probabilistic representation} of user latent variable~(without item). We also provide an alternative interpretation of the Kullback Leibler~(KL) divergence regularization in VAE for CF: we view it as an upper bound of the amount of the information that preserved in the variational distribution, which can allocate proper \textbf{user-level capacity} and avoid over-fitting especially for the sparse signals from inactive users.

Benefit from the probabilistic representations for both the interaction data and auxiliary information, we design a synchronous \textbf{collaborative learning mechanism}: unlike the asynchronous "collaboration" of CDL and CVAE, we adopt KL Divergence to make the probabilistic representation learned from two data views to match with each other at each iteration of the optimization. Compared with previous works, it provides a simple but more effective way to make the information flows between user interaction data and auxiliary user information in bi-direction rather than one-direction. Furthermore, because of the versatility of VAE, the VCM model is not limited to taking the review as the auxiliary information. Different multimedia modalities, e.g., images and other texts, are unified in the framework. Our contribution can be summarized as follows:
\begin{itemize}
\item Unlike previous hybrid models that learns user/item latent factors by attaining maximum a posterior estimates for interaction data, we propose to use two stream VAEs set up to learn the probabilistic representation of user latent variable and provides user-level capacity.
\item Unlike the asynchronous mutual regularization used in previous models, we have the two components learning with each other under a synchronous collaborative learning mechanism, which allows the model to make full use of the Bayesian probabilistic representations from interaction data and auxiliary information.
\item Extensive experiment on three real-world datasets has shown that VCM can significantly outperform the state of the art models. Ablation studies have further proved that improvements come from specific components.
\end{itemize}
\section{Methodology}
Similar to the work in~\cite{Hu:2008:CFI:1510528.1511352}, the recommendation task we processed in this paper accepts implicit feedback. We use a binary matrix $X\in\mathbb{N}^{U\times I}$ to indicate the click~\footnote{we use the verb "click" for concreteness to indicate any interactions, including "check-in," "purchase," "watch"} history among user and item. We use $u\in{1,\ldots,U}$ to indicate users and $i\in{1,\ldots,I}$ to indicate items. The lower case $x_{u}=[x_{u1},\ldots,x_{uI}]\in\mathbb{N^{I}}$ is a binary vector indicating the click history for each item from user $u$. Each user's reviews are merged into one document, let $Y\in\mathbb{N}^{U\times V}$ be the bag-of-words representation for review documents of $U$ users~(where $V$ is the length of the vocabulary). We use $v\in{1,\ldots, V}$ to indicate each word. The lower case $y_{u}=[y_{u1},\ldots,y_{uV}]\in\mathbb{N^{V}}$ is a bag-of-words vector with the number of each word from the document of user $u$.

\subsection{Architecture}
\begin{figure}[t]
\centering
\includegraphics[width=0.45\textwidth]{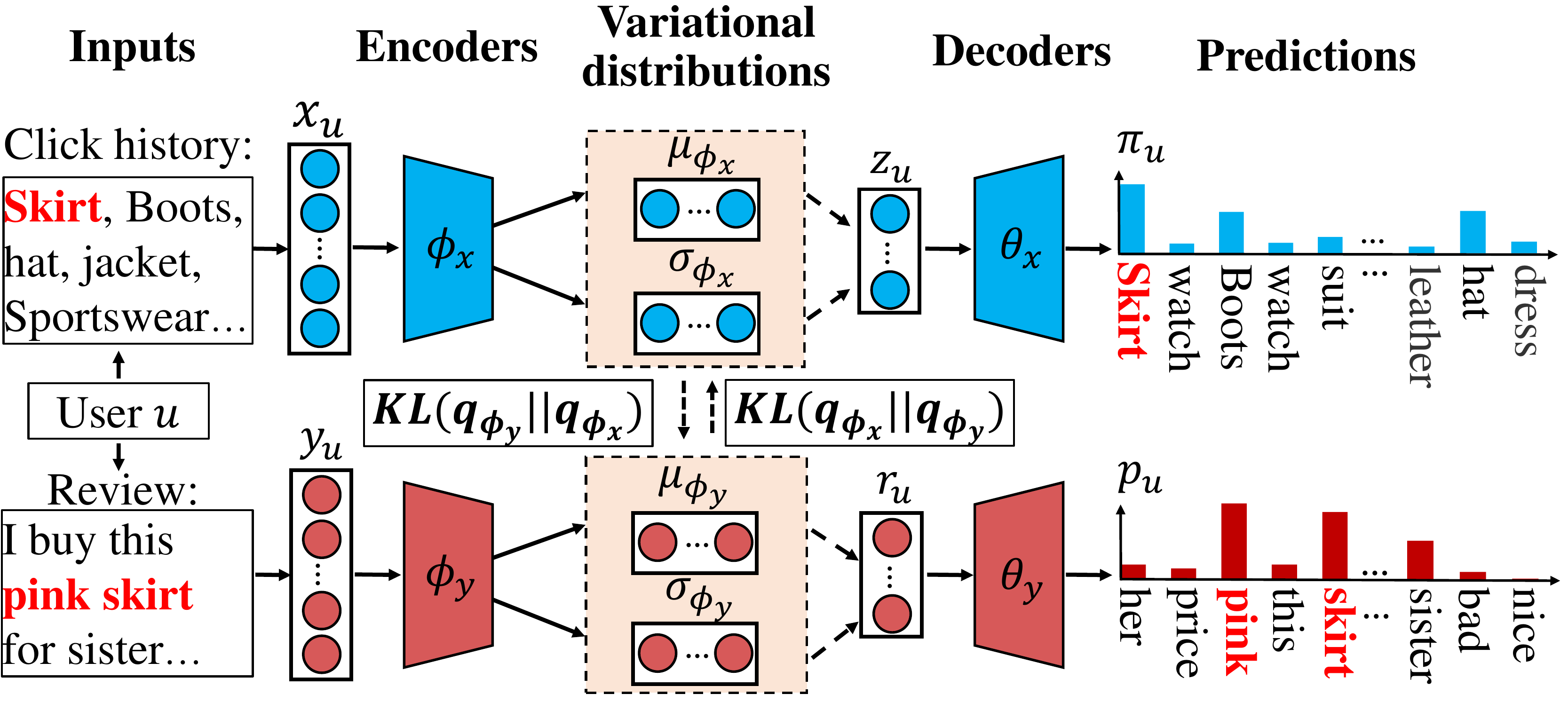}
\caption{VCM model architecture.}
\label{fig::framework}
\end{figure}

The architecture of our proposed model is shown in Figure~\ref{fig::framework}. The model is consists of two parallel extended VAEs, one VAE~($\textrm{VAE}_{x}$) takes users' click history $x_{u}$ as input and output the probability over items, one VAE~($\textrm{VAE}_{y}$) takes users' review text data $y_{u}$ as input and output the probability over words. Each VAE uses the \textbf{encoder} to compresses the input to the variational distribution then transfers the latent variable sampled from the posterior to the \textbf{decoder} to get the generative distribution for prediction. The KL divergence between two variational distributions is employed for the cooperation between $\textrm{VAE}_{x}$ and $\textrm{VAE}_{y}$.

\subsubsection{Encoders}
We assume that the interaction data click history $x_{u}$ can be generated by user latent variable $z_{u}\in \mathbb{R}^{K}$, and the auxiliary information review document $y_{u}$ can be generated by the another user latent variable $r_{u} \in \mathbb{R}^{K}$. We introduce the variational distribution $q_{\phi_{x}}(z_{u}|x_{u})$ and $q_{\phi_{y}}(r_{u}|y_{u})$ to approach the true posteriors $p(z_{u}|x_{u})$ and $p(r_{u}|y_{u})$, which represent the user click behavior preference and review document semantic content, respectively. Here, we employ the parameterised diagonal Gaussian $\mathcal{N}(\mu_{\phi_{x}}, \textrm{diag}\{\sigma^{2}_{\phi_{x}}\})$ as $q_{\phi_{x}}(z_{u}|x_{u})$, and employ $\mathcal{N}(\mu_{\phi_{y}},\textrm{diag}\{\sigma^{2}_{\phi_{y}}\})$ as $q_{\phi_{y}}(r_{u}|y_{u})$. So we define the inference process of the probabilistic encoders as below:
\begin{enumerate}
	\item Construct vector representations of observed data for user $u$:
\begin{displaymath}
\begin{aligned}
j_{u}=f_{\phi_{x}}^{\textrm{DNN}}(x_{u}),~e_{u}&=f_{\phi_{y}}^{\textrm{DNN}}(y_{u}).  
\end{aligned}
\end{displaymath}
	\item Parameterise the variational distribution over the user latent variables $r_{u}$ and $z_{u}$:
\begin{displaymath}
\begin{split}
[\mu_{\phi_{x}}(x_{u}),\sigma_{\phi_{x}}(x_{u})] = l_{\phi_{x}}(j_{u}) \in \mathbb{R}^{2K},\\
[\mu_{\phi_{y}}(y_{u}),\sigma_{\phi_{y}}(y_{u})] = l_{\phi_{y}}(e_{u}) \in \mathbb{R}^{2K}.
\end{split}
\end{displaymath}
\end{enumerate}
$f_{\phi_{x}}^{\textrm{DNN}}(\cdot)$ and $f_{\phi_{y}}^{\textrm{DNN}}(\cdot)$ can be any type of deep neural networks~(DNN) that are suitable for the observed data. $l_{\phi_{x}}(\cdot)$ and $l_{\phi_{y}}(\cdot)$ are linear transformation, computing the parameters of the variational distributions. And $\phi_{x}$ is consist of the parameters of $f_{\phi_{x}}^{\textrm{DNN}}(\cdot)$ and $l_{\phi_{x}}$, whereas $\phi_{y}$ is consist of the parameters of $f_{\phi_{y}}^{\textrm{DNN}}(\cdot)$ and $l_{\phi_{y}}$.

\subsubsection{Decoders}
 We define the generation process of two softmax decoders as below:
\begin{enumerate}
\item Draw samples $z_{u} \in \mathbb{R}^{K}$ and $r_{u} \in \mathbb{R}^{K}$ from variational posterior $q_{\phi_{x}}(z_{u}|x_{u})$ and $q_{\phi_{y}}(r_{u}|y_{u})$, respectively.
\item Produce the probabilistic distribution over $I$ items and $V$ words for each user through DNN and softmax function:
\begin{displaymath}
\begin{aligned}
\pi_{ui} = \frac{\textrm{exp}(f_{\theta_{x}}^{\textrm{DNN}}(z_{u})_{i})}{\sum_{i}^{I}{\textrm{exp}(f_{\theta_{x}}^{\textrm{DNN}}(z_{u})_{i})}}~,
p_{uv} = \frac{\textrm{exp}(f_{\theta_{y}}^{\textrm{DNN}}(r_{u})_{v})}{\sum_{v}^{V}{\textrm{exp}(f_{\theta_{y}}^{\textrm{DNN}}(r_{u})_{v})}}.
\end{aligned}
\end{displaymath}
\item Reconstruct the data from two multinomial distributions, respectively:
\begin{displaymath}
\begin{aligned}
x_{u} \sim \textrm{Mult}(N_{u},\pi_{u}),~y_{u} \sim \textrm{Mult}(W_{u}, p_{u}).
\end{aligned}
\end{displaymath}
\end{enumerate}
where $f_{\theta_{x}}^{\textrm{DNN}}$ and $f_{\theta_{y}}^{\textrm{DNN}}$ are two DNN with parameters $\theta_{x}$ and $\theta_{y}$. $N_{u} = \sum_{i}^{I}{x_{ui}}$ is the sum of clicks, and $W_{u} = \sum_{v}^{V}{y_{uv}}$ is the sum of words in review document of user $u$, the observed data $x_{u}$ and $y_{u}$ can be generated from the two multinomial distribution respectively.
Therefore, a suitable goal for learning the distribution of latent variable $z_{u}$ is to maximize the marginal log-likelihood function of click behavior data in expectation over the whole distribution of $z_{u}$,
\begin{displaymath}
\begin{aligned}
&\mathop{\max}_{\theta_{x},\phi_{x}}\mathbb{E}_{q_{\phi_{x}}(z_{u}|x_{u})}[\textrm{log}~ p_{\theta_{x}}(x_{u}|z_{u})],\\
&\textrm{log}~ p_{\theta_{x}}(x_{u}|z_{u})=\sum\nolimits_{i}^{I}{x_{ui} \textrm{log}~\pi_{ui}}.
\end{aligned}
\label{equ::max_maginal_likelihood}
\end{displaymath}
And we can also get similar likelihood function of review document, we omitted the similar process for space limitation. 

\subsubsection{User-level Capacity}

We introduce a limitation over $q_{\phi}(z_{u}|x_{u})$ to control the capacity of different users. This can be achieved if we match $q_{\phi}(z_{u}|x_{u})$ with the uninformative prior, such as the isotropic unit Gaussian used in \cite{higgins2016early,higgins2016beta}. Hence, we get the constrained optimization problem for the marginal log-likelihood function of click behavior data as:
\begin{displaymath}
\begin{aligned}
&\mathop{\max}_{\theta_{x},\phi_{x}} \mathbb{E}_{q_{\phi}(z_{u}|x_{u})}[\textrm{log}~ p_{\theta_{x}} 
(x_{u}|z_{u})],\\
&\textrm{subject to}~{KL}(q_{\phi}(z_{u}|x_{u})||p(z_{u} )) < c_{u},
\end{aligned}
\label{equ::constrain_problem_for_click}
\end{displaymath}
${KL}(q_{\phi_{x}}(z_{u}|x_{u})||p(z_{u}))$ has the property of being zero if the posterior distribution is equal to the uninformative prior, which means the model learn nothing from the data. Thus, the hidden variable $c_{u}$ can be seen as the upper bound of the amount of information that preserved in the variational distribution for each user's preference. According to complementary slackness KKT conditions~\cite{kuhn1951aw,karush1939minima}, solving this optimization problem is equivalent to maximize the lower bound as below:
\begin{displaymath}
\begin{aligned}
&\mathcal{L}_{x}(\theta_{x},\phi_{x};x_{u},z_{u},\beta_{x}) \\
=&~\underbrace{\mathbb{E}_{q_{\phi_{x}}(z_{u}|x_{u})}[\textrm{log}~ p_{\theta_{x}}(x_{u}|z_{u})]}_{\textrm{Reconstruction loss}}-\beta_{x}\underbrace{{KL}(q_{\phi_{x}}(z_{u}|x_{u})||p(z_{u}))}_{\textrm{Capacity limitation regularization}}
\end{aligned}
\end{displaymath}
So far, we get the lower bound $\mathcal{L}_{x}$ for $\textrm{VAE}_\textrm{x}$, similar process can be done to obtain the lower bound $\mathcal{L}_{y}$ for $\textrm{VAE}_\textrm{y}$ as:
\begin{displaymath}
\begin{aligned}
 &\mathcal{L}_{y}(\theta_{y},\phi_{y};y_{u},r_{u},\beta_{y}) \\
=&~\mathbb{E}_{q_{\phi_{y}}(r_{u}|y_{u})}[\textrm{log}~ p_{\theta_{y}}(y_{u}|r_{u})]-\beta_{y}{KL}(q_{\phi_{y}}(r_{u}|y_{u})||p(r_{u}))
\end{aligned}
\end{displaymath}

Varying KKT multiplier $\beta_{x}$, $\beta_{y}$ puts different strength into pushing the variational distribution to align with the unit Gaussian prior. A proper choice of $\beta_{x}$, $\beta_{y}$ can balance the trade-off between reconstruction loss and the limitation.

\subsubsection{Collaborative Learning Mechanism}

To improve the generalization recommendation performance of variational CF model $\textrm{VAE}_\textrm{x}$, we use $\textrm{VAE}_\textrm{y}$ as the teacher to provide review semantic content in the form of the posterior probability $q_{\phi_{y}}$ to guide the learning process of $\textrm{VAE}_\textrm{x}$. To measure the match of two posterior distributions $q_{\phi_{x}}$ and $q_{\phi_{y}}$, we adopt KL divergence. The KL distance from $q_{\phi_{y}}$ to $q_{\phi_{x}}$ is computed as:
\begin{displaymath}
\begin{aligned}
{KL}(q_{\phi_{x}}(z_{u}|x_{u})||q_{\phi_{y}}(r_{u}|y_{u})).
\label{equ::KKT_click}
\end{aligned}
\end{displaymath}
Similarly, to improve the ability to learn representation of semantic meaning for $\textrm{VAE}_\textrm{y}$, we use $\textrm{VAE}_\textrm{x}$ as teacher to provide click behavior preference information in form of its posterior $q_{\phi_{x}}$ to guide the $\textrm{VAE}_\textrm{y}$ to capture the semantic content for review document, so the KL distance from $q_{\phi_{x}}$ to $q_{\phi_{y}}$ is computed as:
\begin{displaymath}
\begin{aligned}
{KL}(q_{\phi_{y}}(r_{u}|y_{u})||q_{\phi_{x}}(z_{u}|x_{u})).
\label{equ::KKT_click}
\end{aligned}
\end{displaymath}

We adopt this bi-directional KL Divergence to make the probabilistic representation learned from two data views to match itself with each other, so that allows the VCM to fully leverage the two probabilistic representation. 

\subsection{Objective Function}
We form the objective for user $u$ with collaborative learning mechanism as~(we can get the objective function of the dataset by averaging the objective function for all users):
\begin{displaymath}
\begin{aligned}
&\mathcal{L}(\phi,\theta;x_{u},z_{u},y_{u},r_{u},\beta_{x},\beta_{y}) \\
=&\mathcal{L}_{x}(\theta_{x},\phi_{x};x_{u},z_{u},\beta_{x}) - \beta_{x}{KL}(q_{\phi_{x}}(z_{u}|x_{u})||q_{\phi_{y}}(r_{u}|y_{u})) \\
+&\mathcal{L}_{y}(\theta_{y},\phi_{y};y_{u},r_{u},\beta_{y}) - \beta_{y}{KL}(q_{\phi_{y}}(r_{u}|y_{u})||q_{\phi_{x}}(z_{u}|x_{u}))
\label{equ::object_click}
\end{aligned}
\end{displaymath}
Note the parameters need optimize is $\phi = \{\phi_{x},\phi_{y}\}$,$\theta=\{\theta_{x}, \theta_{y}\}$. We can obtain an unbiased estimate of $\mathcal{L}$ by sampling $z_{u}\sim q_{\phi_{x}}$, and $r_{u}\sim q_{\phi_{y}}$, then perform stochastic gradient ascent to optimize it. And by doing \textit{reparameterization trick}~\cite{kingma2013auto}: we sample $\varepsilon \sim \mathcal{N}(0,I_{K})$, and reparameterize $z_{u}=\mu_{\phi_{x}}(x_{u}) + \varepsilon \odot \sigma_{\phi_{x}}(x_{u})$, $r_{u}=\mu_{\phi_{y}}(y_{u}) + \varepsilon \odot \sigma_{\phi_{y}}(y_{u})$, the stochasticity of the sampling process is isolated, the gradient with respect to $\phi$ can be back-propagated through the sampled $z_{u}$ and $r_{u}$. With $\mathcal{L}$ as the final lower bound, we train this two VAEs synchronously at each iteration according to Algorithm~\ref{alg:1}.
\begin{algorithm}[t]
	\renewcommand{\algorithmicrequire}{\textbf{Input:}}
	\caption{VCM collaborative training with anneal stochastic gradient descent}
	\label{alg:1}
	\begin{algorithmic}[1]
		\REQUIRE Click matrix $X\in\mathbb{N}^{U\times I}$, Bag of word representation of review $Y\in\mathbb{N}^{U\times V}$, $\beta$, Anneal steps
        \STATE Randomly initialize $\phi$, $\theta$
        \FOR {iteration \textbf{in} Anneal steps}
		\STATE Sample a batch of users $\mathcal{U}$
		\FORALL{$u \in \mathcal{U}$}
		\STATE Compute $z_{u}$ and $r_{u}$ via reparameterization trick
        \STATE Compute noisy gradient $\nabla_{\phi}{\mathcal{L}}$, $\nabla_{\theta}{\mathcal{L}}$ with $z_{u}$ and $r_{u}$
		\ENDFOR
        \STATE Average noisy gradient from batch
        \STATE $\beta_{x} = \beta_{y}= \textrm{min}(\beta,\textrm{iteration}/\textrm{Anneal steps})$
        \STATE Update $\phi$ and $\theta$ by taking gradient update with $\beta_{x}$, $\beta_{y}$
        \ENDFOR
		\STATE \textbf{return} $\phi$, $\theta$
	\end{algorithmic}  
\end{algorithm}
\subsection{Prediction}
We now describe how we make predictions given a trained model. Given a user's click history $x_{u}$, we rank all the items based on the predicted multinomial probability $\pi_{u}$. The latent variables $z_{u}$ for $x_{u}$ is constructed as follows: we simply take the mean of the variational distribution $z_{u}=\mu_{\phi_{x}}(x_{u})$. We denote this prediction method as \textbf{VCM}.

Benefit from collaborative learning, our model allows for bi-directional prediction (review2click and click2review). In order to predict click behavior corresponding to user's review semantic content, we infer the latent variables $r_{u}$ by presenting the reviews $y_{u}$ to encoder of $\textrm{VAE}_{\textrm{y}}$, we also simply take the mean $r_{u}=\mu_{\phi_{y}}(y_{u})$ to construct the latent variable $r_{u}$ and use the decoder of $\textrm{VAE}_{\textrm{x}}$ with $r_{u}$ as input to generate the predicted multinomial probability $\pi_{u}$. So now only given a user's review document, our model can encode the text into the latent variables and decode it to click behavior. And we denote this \textbf{C}ross-\textbf{D}omain prediction method as \textbf{VCM-CD}.

\section{Experiment}
\subsection{Datasets}
We experimented with three publicly accessible datasets from various domains with different scale and sparsity.
\begin{itemize}
    \item \textbf{Yelp-2013}~(Yelp): This data~\cite{seo2017interpretable} contains user-business check-in record and reviews from \textit{RecSys Challenge 2013}~\footnote{https://www.kaggle.com/c/yelp-recsys-2013/data}. We only keep users who have checked in at least five business.
 \item \textbf{Amazon Clothing}~(Clothing): The \textit{Amazon} dataset is the consumption records with reviews from \textit{Amazon.com}. We use the \textit{the clothing shoes and jewelry category 5-core}~~\cite{He:2016:VVB:3015812.3015834}. We only keep users with at least five products in their shopping record and products that are bought by at least 5 users.
\item \textbf{Amazon Movies}~(Movies): This data ~\cite{He:2016:VVB:3015812.3015834} contains the user-movie rating from \textit{Movies and TV 5-core} with reviews. We only keep user with 5 watching record and movies that are played at least 10 users.
\end{itemize}
For each data set, we binaries the explicit data by maintaining ratings of four or higher and interpret it as implicit feedback. We merge each user's reviews into one document, then we follow the same process to remove the stop words as~\cite{miao2016neural} for each document, and keep the most common $V=10,000$ words in all documents as the vocabulary. Table~\ref{Tab::dataset} summarizes the characteristics of all the datasets after pre-processing.
\begin{table}[]
\caption{Statistical characteristics of the datasets after preprocessing}
\begin{tabular}{|l|c|c|c|}
\hline
                   & Yelp   & Clothing & Movies \\ \hline
\# of users        & 6,784   & 21,181        & 81,780        \\\hline
\# of items        & 10,003  & 17,710        & 24,628        \\\hline
\# of interactions & 106,630 & 145,281       & 1,028,839      \\\hline

sparsity & 0.16\% & 0.04\%       & 0.05\%      \\ \hline

\end{tabular}
\label{Tab::dataset}
\end{table}

\subsection{Metric}
We use two ranking-based metrics: the truncated normalized discounted cumulative gain~(NDCG@$R$) and Recall@$R$. For each user, both the metrics compare the predicted rank of the held-out items with their true rank. Moreover, we get the predicted rank by sorting the multinomial probability $\pi_{u}$. Formally, we define $w(r)$ as the item at rank $r$, $\mathbb{I}[\cdot]$ is the indicator function, and $I_{u}$ is the set of the held-out items that user $u$ clicked on.
\begin{displaymath}
\begin{aligned}
\textrm{Recall}@R(u,w) = \frac{\sum_{r=1}^{R}{\mathbb{I}[w(r)\in I_{u}]}}{min(R,I_{u})}
\end{aligned}
\end{displaymath}

The expression in the denominator is the minimum of the number of items clicked by user $u$ and $R$. While Recall@$R$ considers all items ranked with the first $R$ to be equally important, and it reaches to the maximum of 1 when the model ranks all relevant items in position.
And the truncated discounted cumulative gain~(DCG@$R$) is
\begin{displaymath}
\begin{aligned}
\textrm{DCG}@R(u,w) = \sum_{r=1}^{R}{\frac{2^{\mathbb{I}[w(r)\in I_{u}]}-1}{\textrm{log}(r+1)}}
\end{aligned}
\end{displaymath}
DCG@$R$ assign higher scores to the higher ranks versus lower ones. NDCG@$R$ is the DCG@$R$ linearly normalized to $[0,1]$ after dividing by the best possible DCG@$R$ when all the held-out items are ranked at the top. 
\subsection{Baselines}
As the previous works~\cite{Wang:2015:CDL:2783258.2783273,Li:2017:CVA:3097983.3098077,Zheng:2017:JDM:3018661.3018665} has demonstrated, the performance of hybrid recommendation with auxiliary information is significantly better than CF-based models, so only hybrid models are used for comparison. The baselines included in our comparison are as follows:

\begin{itemize}
\item \textbf{CDL}: \textbf{C}ollaborative \textbf{D}eep \textbf{L}earning~\cite{Wang:2015:CDL:2783258.2783273} tightly combines the SDAE with the PMF. The middle layer of the neural network acts as a bridge between the SDAE and the PMF.

\item \textbf{CVAE}: \textbf{C}ollaborative \textbf{V}ariational\textbf{A}utoencoder~\cite{Li:2017:CVA:3097983.3098077} is a probabilistic feedforward model for joint learning of VAE and collaborative filtering. CVAE is a very strong baseline and achieves the best performance among our baseline methods.

\item \textbf{DeepCoNN}: \textbf{D}eep \textbf{C}ooperative \textbf{N}eural \textbf{N}etworks~\cite{Zheng:2017:JDM:3018661.3018665} jointly models user and item from textual reviews for rating prediction. To make it comparable, we revise the model to suitable for implicit feed back with negative sampling~\cite{He:2017:NCF:3038912.3052569}.

\end{itemize}

\subsection{Experimental setup}

We randomly split the interaction data into training, validation, test sets. For each user, we take $60\%$ of the entire click history as $x_{u}$ and review document as $y_{u}$ to train models.

For evaluation, we use 20\% click history as the validation set to tune hyper-parameters, and 20\% held-out click history as the test set. We can take click history in train set~(VCM prediction) or the review document~(VCM-CD prediction) to learn the necessary users' representations, and then compute metrics by looking how well the model ranks the rest unseen click items from the held-out set.

We select models hyper-parameters and architectures by evaluating NDCG@100 on the validation sets. For VCM, we explore Multilayer perceptron~(MLP) with 0,1 and 2 hidden layers, and we find the best overall architecture for VCM would be $[I \to 600 \to K \to 600 \to I]$ for $\textrm{VAE}_{\textrm{x}}$ and $[V \to 500 \to K \to V]$ for $\textrm{VAE}_{\textrm{y}}$. Moreover, we find that going deeper does not improve performance. We use tanh as the activation function between layers. Note that since the output of $l_{\phi_{x}}$ and $l_{\phi_{y}}$ are used as the mean and variance of the Gaussian random variables, we do not apply an activation function on it. We apply dropout at the input layer with probability $0.5$ for $\textrm{VAE}_{\textrm{x}}$. We do not apply the weight decay for any parts. We train our model using Adam~\cite{DBLP:journals/corr/KingmaB14} with the batch size of $128$ users for 200 epoch on both datasets. We save the model with the best validation NDCG@$100$ and report test set metrics with it. For simplicity, we set $\beta_{x}$ and $\beta_{y}$ with the same value and anneal them linearly for $40,000$ anneal steps, using the schedule described in Algorithm~\ref{alg:1}.

\begin{figure}[htbp]
\centering
\begin{minipage}[t]{0.23\textwidth}
\centering
\includegraphics[width=\textwidth]{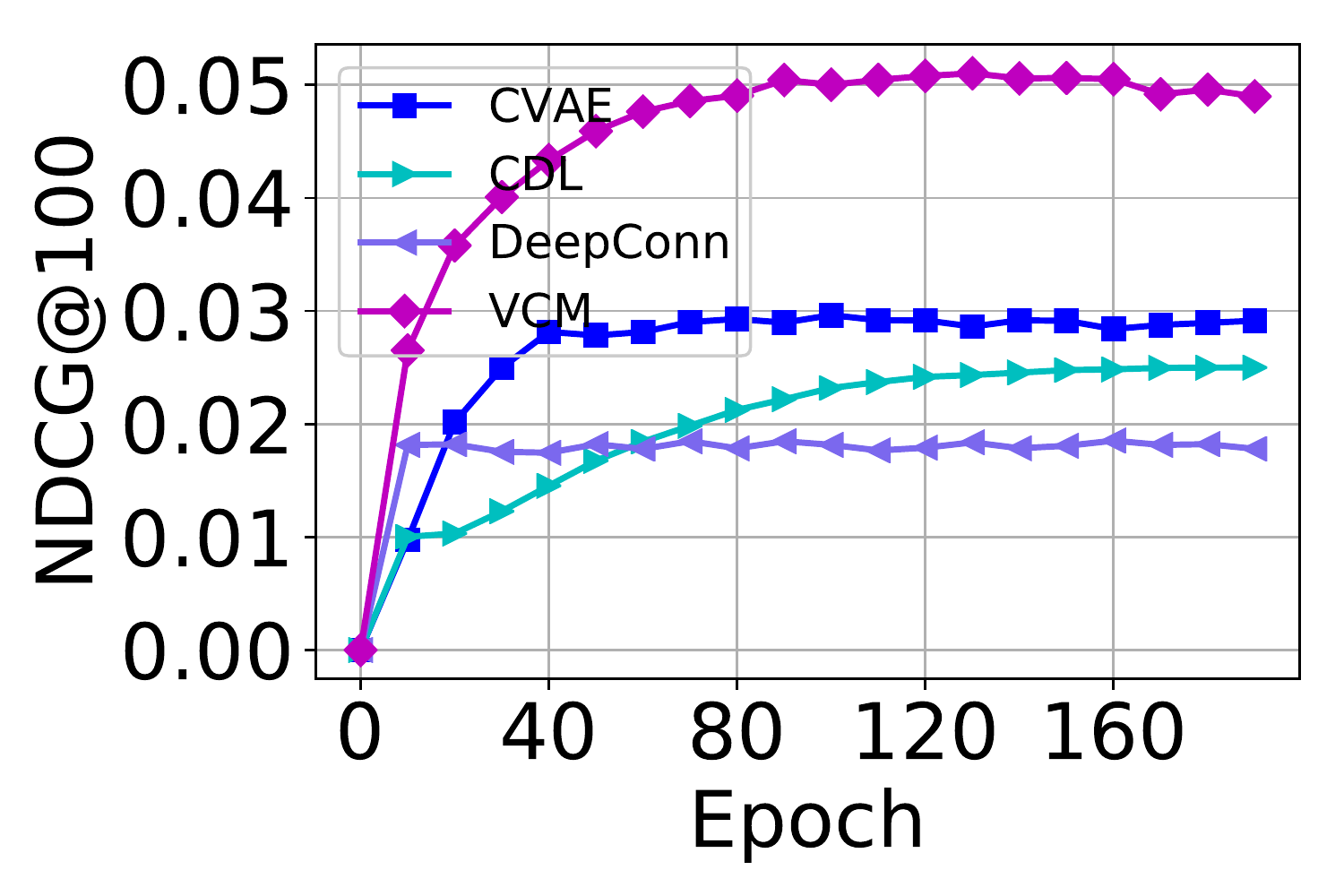}
\caption{Performance of NDCG@100 of all models w.r.t. the number of the epoch on Clothing validation set}
\label{fig::NDCG_on_validation_set_in_training}
\end{minipage}
\begin{minipage}[t]{0.23\textwidth}
\centering
\includegraphics[width=\textwidth]{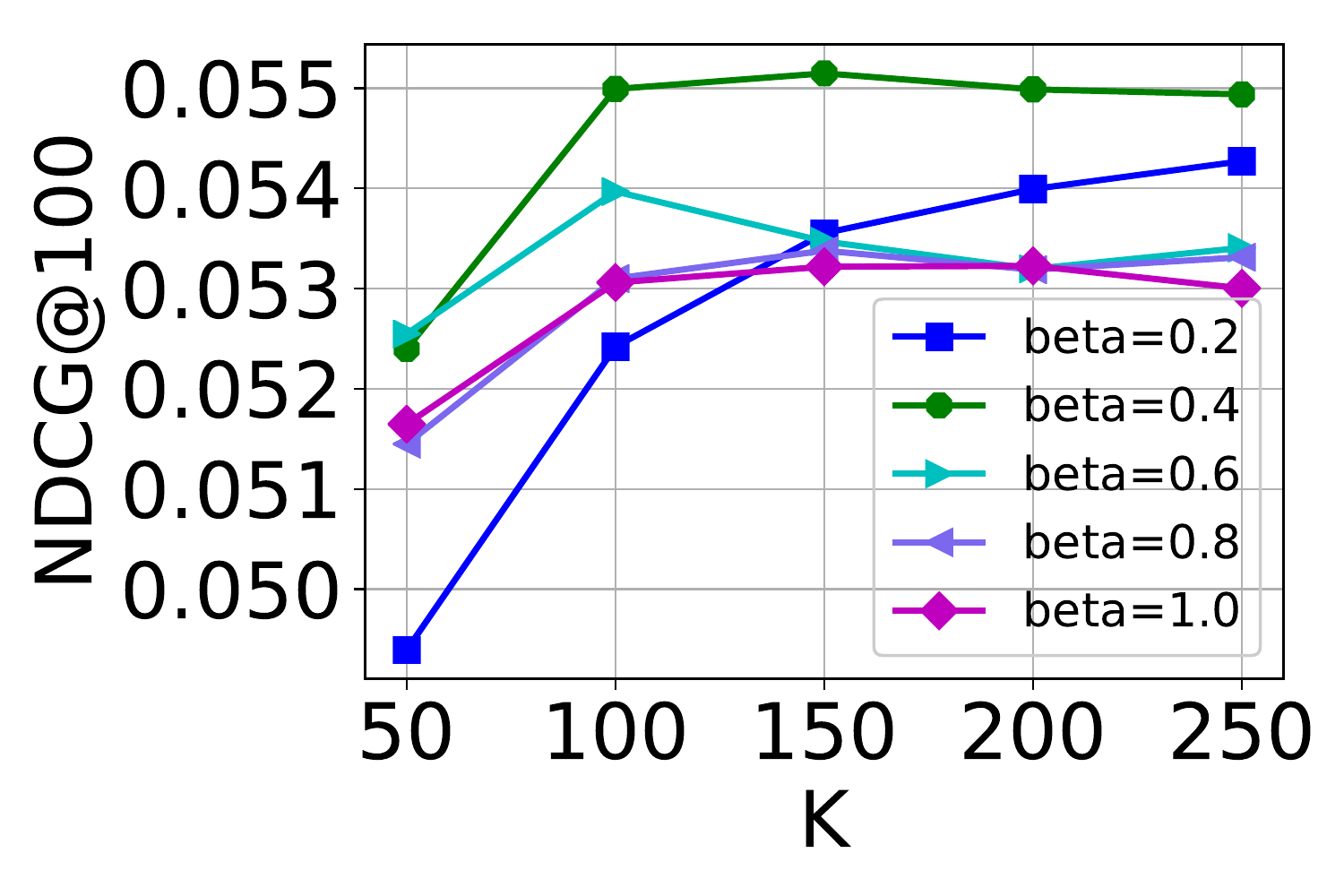}
\caption{Performance of NDCG@100 of VCM w.r.t. the latent dimension and $\beta$ on Clothing validation set}
\label{fig::sensitive to capacity parameters}
\end{minipage}
\end{figure}

Figure~\ref{fig::NDCG_on_validation_set_in_training} shows the NDCG@100 on Clothing validation set during training. Also, we empirically studied the effects of two important parameters of VCM: the latent dimension, the regularization coefficient $\beta_{x}$ and $\beta_{y}$. Figure~\ref{fig::sensitive to capacity parameters} shows the performance of VCM on the validation set of Clothing with varying $K$ from $50$ to $250$ and $\beta_{x}$, $\beta_{y}$ from $0.2$ to $1.0$ to investigate its sensitivity. As it can be seen, the best regularization coefficient is $0.4$, and it does not improve the performance when the dimension of the latent space is greater than 100 in $\beta=0.4$. Results on Movies and Yelp show the same trend, and thus they are omitted due to the space limitation.

\subsection{Quantitative result}
\begin{figure}[ht]
\centering
	\begin{subfigure}[b]{.23\textwidth}
	\centering
    \includegraphics[width=.99\textwidth]{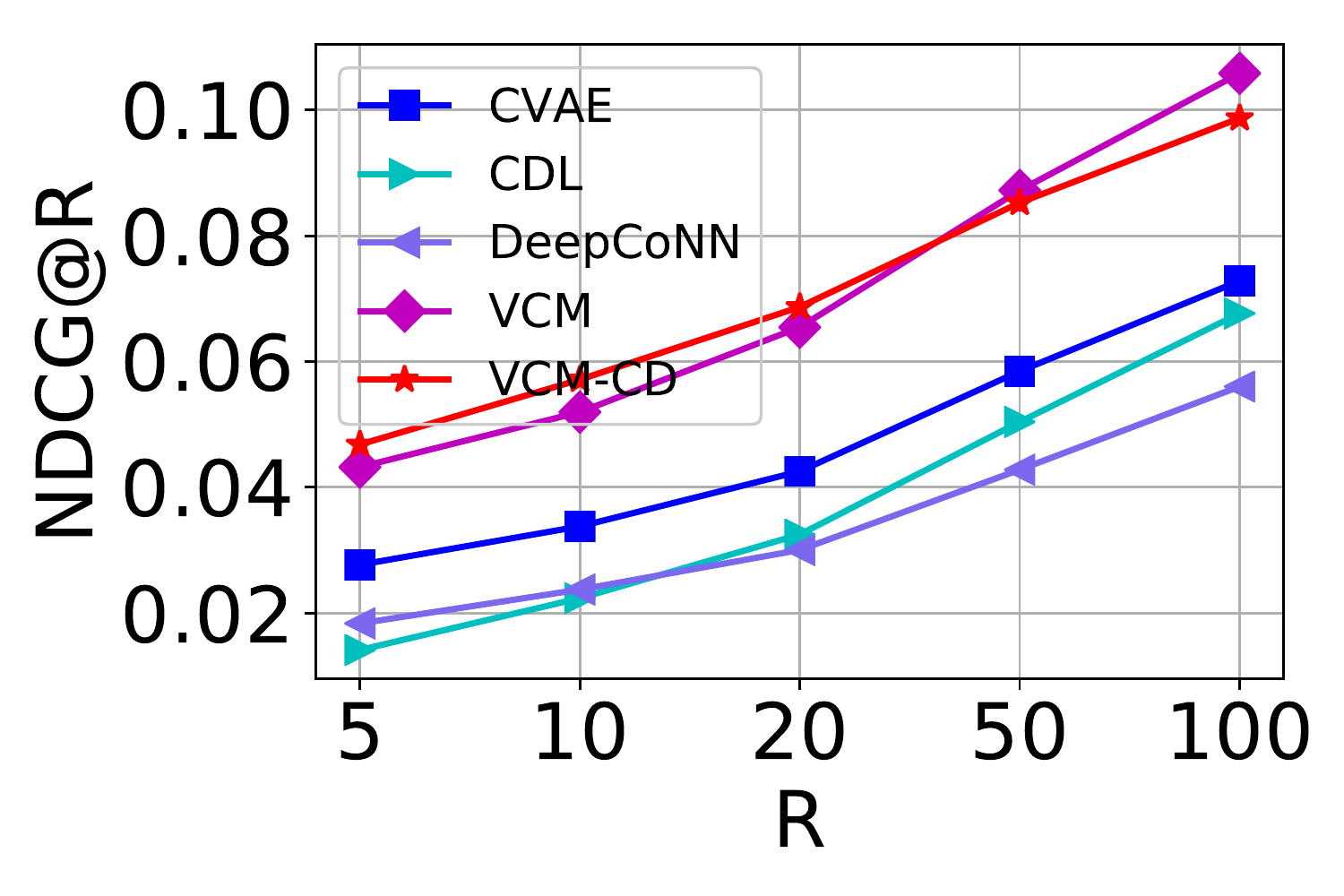}
    \caption{Yelp-NDCG@$R\dag$}
	\end{subfigure}
    	\begin{subfigure}[b]{.23\textwidth}
	\centering
    \includegraphics[width=.99\textwidth]{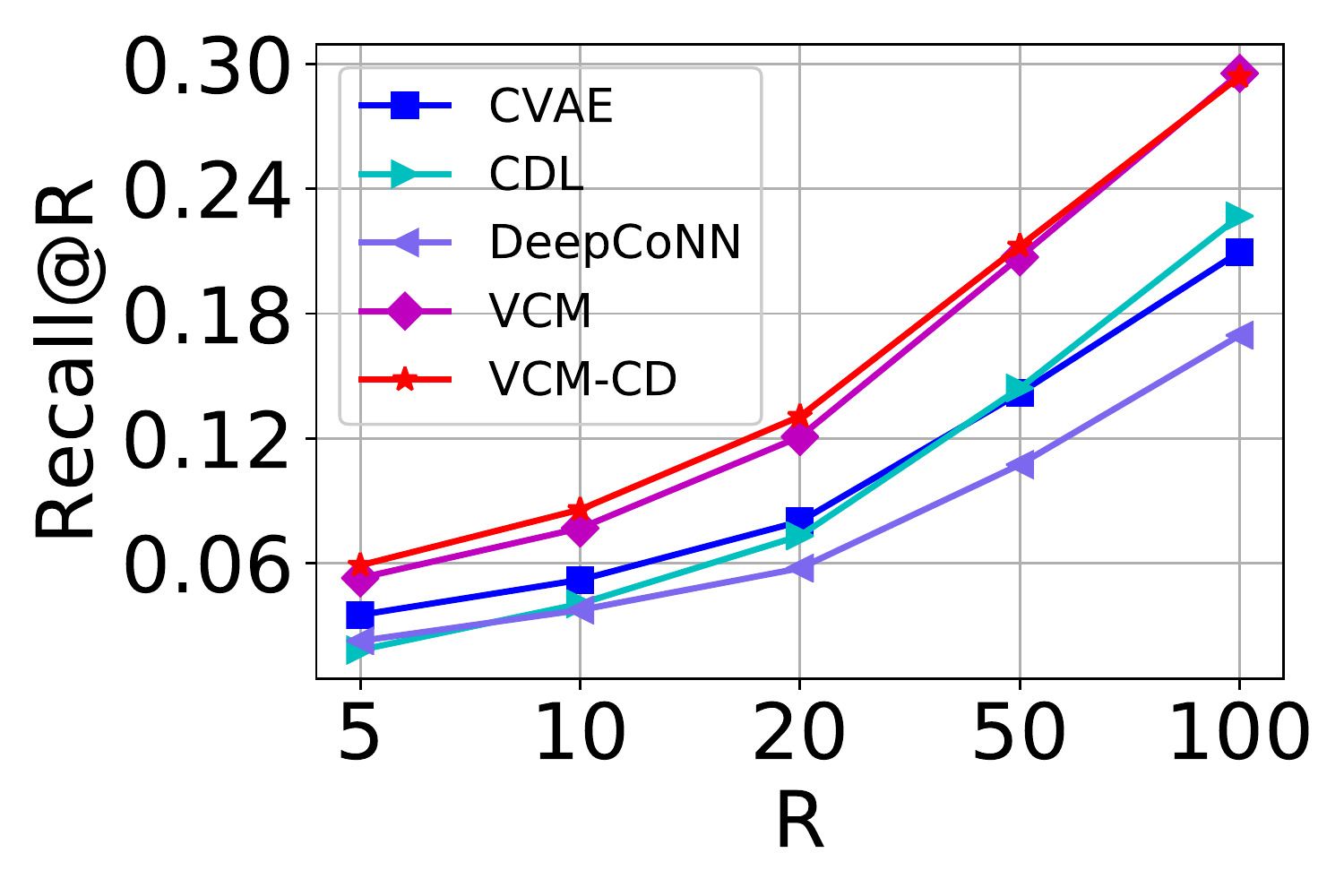}
    \caption{Yelp-Recall@$R\dag$}
	\end{subfigure}

	\begin{subfigure}[b]{.23\textwidth}
	\centering
    \includegraphics[width=.99\textwidth]{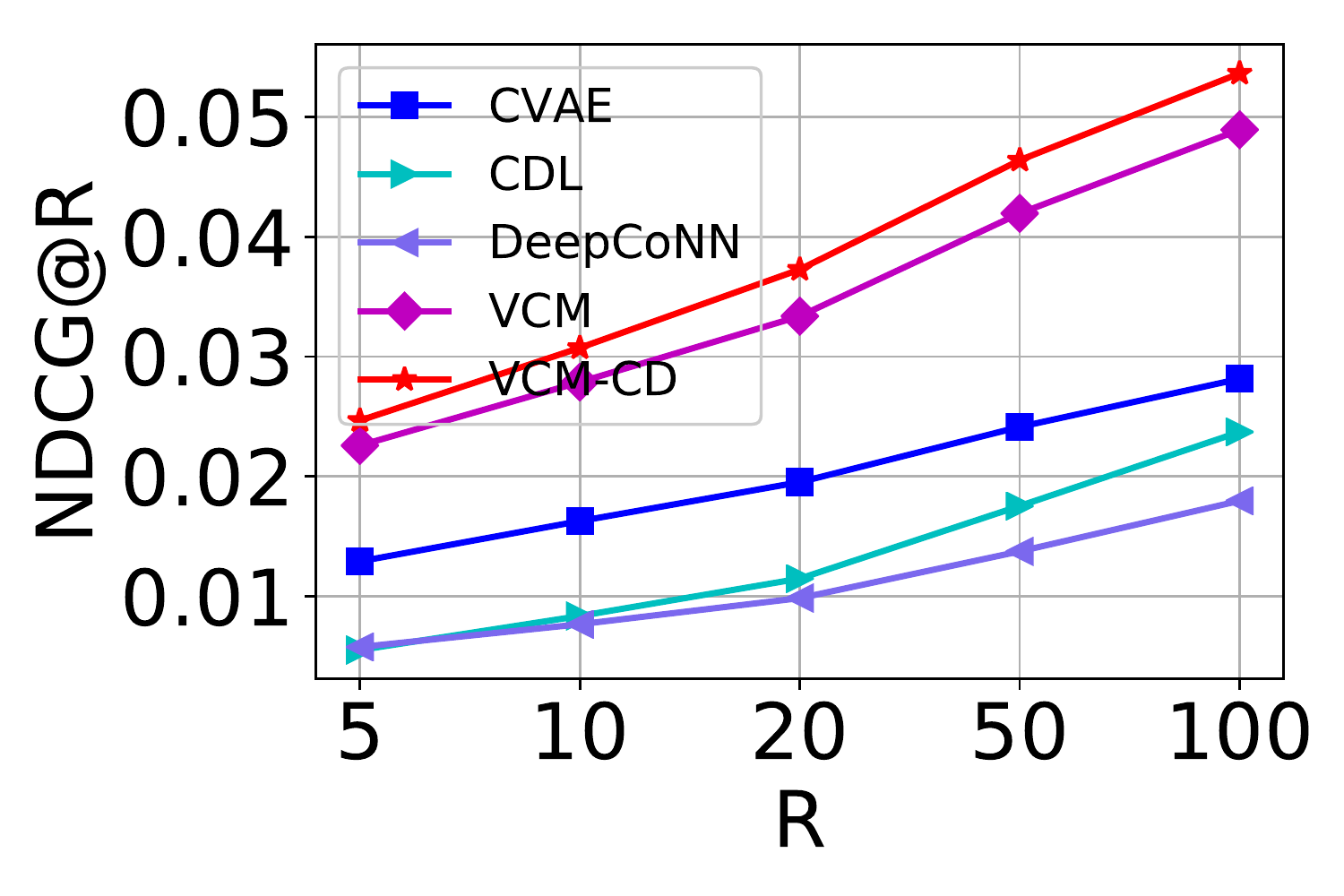}
    \caption{Clothing-NDCG@$R\dag$}
	\end{subfigure}
    	\begin{subfigure}[b]{.23\textwidth}
	\centering
    \includegraphics[width=.99\textwidth]{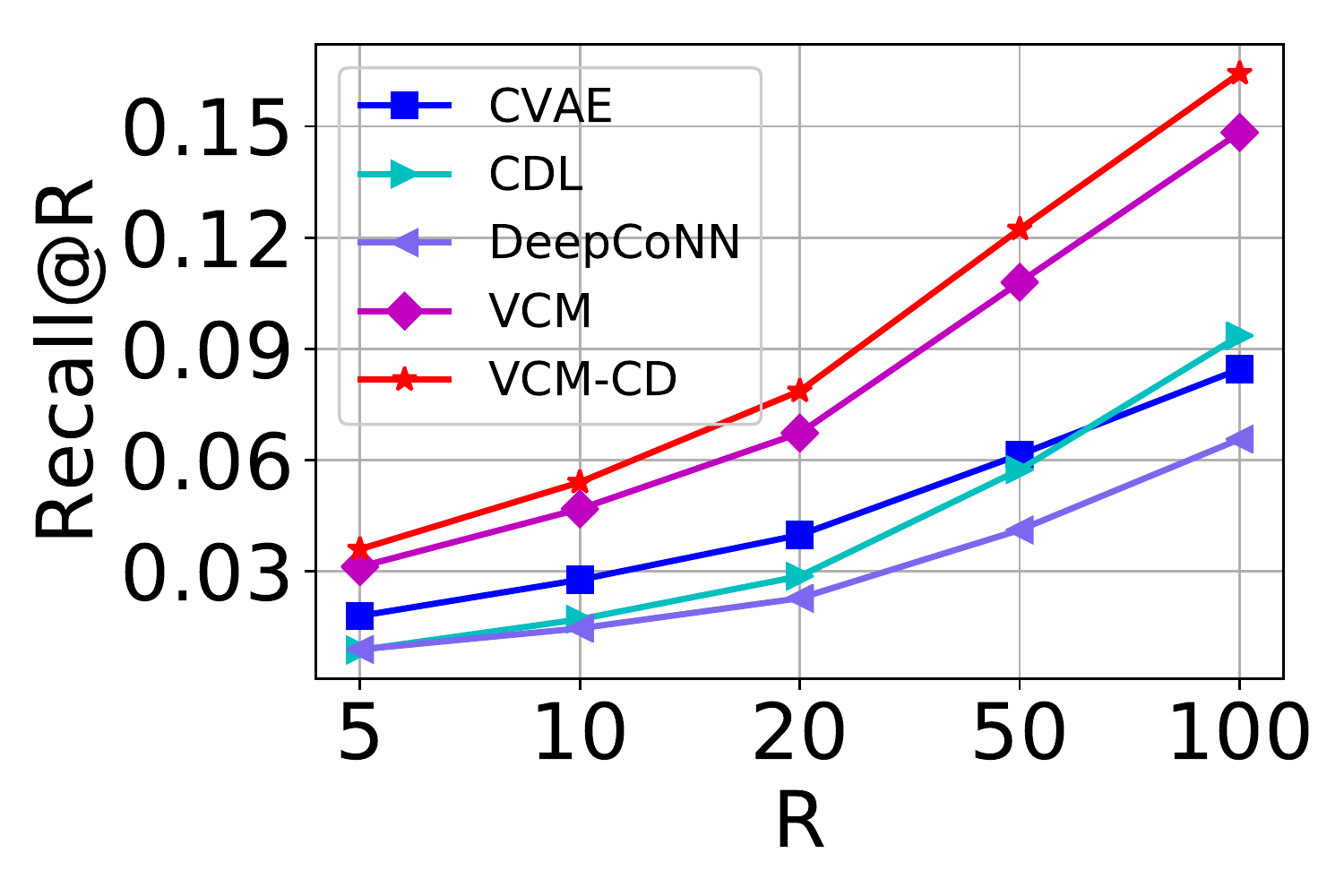}
    \caption{Clothing-Recall@$R\dag$}
	\end{subfigure}

	\begin{subfigure}[b]{.23\textwidth}
	\centering
    \includegraphics[width=.99\textwidth]{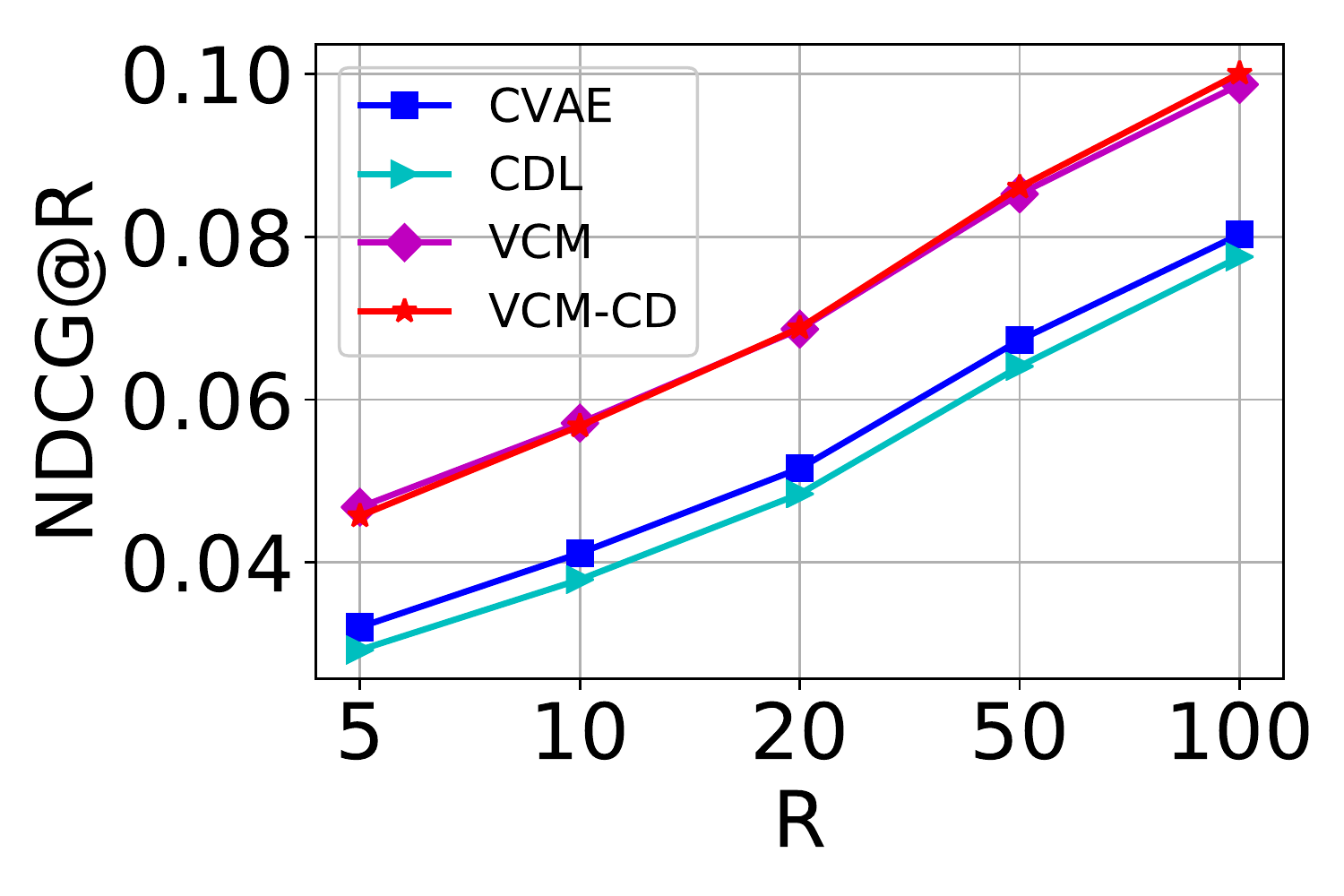}
    \caption{Movies-NDCG@$R\dag$}
	\end{subfigure}
    	\begin{subfigure}[b]{.23\textwidth}
	\centering
    \includegraphics[width=.99\textwidth]{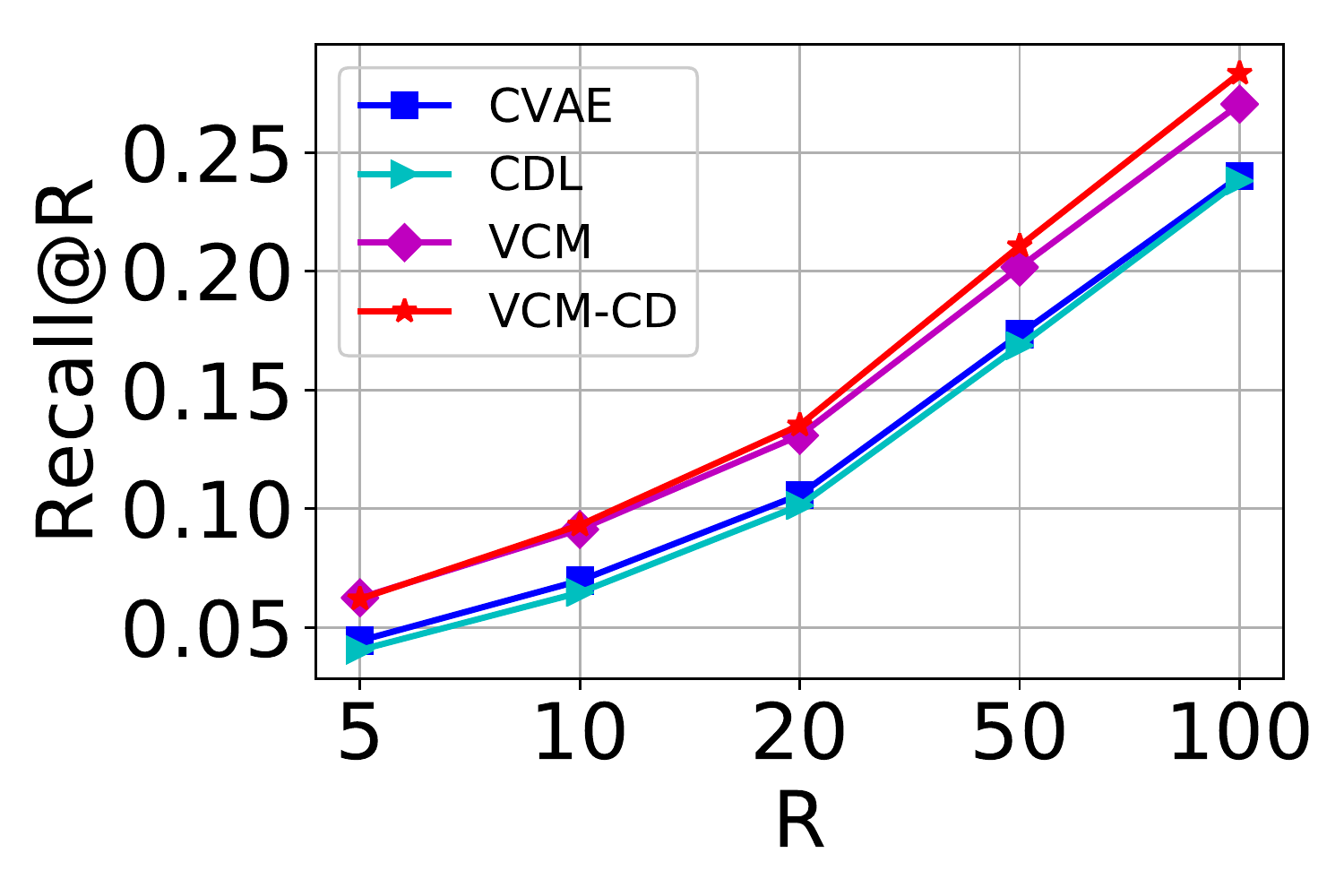}
    \caption{Movies-Recall@$R\dag$}
	\end{subfigure}
    \caption{Evaluation of Top-$R$ item recommendation on three datasets. Standard errors of NDCG@100 are around 8e-4 for Yelp and 4e-4 for Clothing and 2e-4 for Movies. For each subplot, a paired t-test is performed, and $\dag$ indicates statistical significance at $p<0.01$, compared to the best baseline. We could not finish DeepCoNN within a reasonable amount of time on Movies.}\label{fig::result}
\end{figure}

Figure~\ref{fig::result} summarizes the results between our proposed methods and various baselines. The experiments are repeated 10 times, and the averages are reported. Each metric averaged across all users. Both VCM and VCM-CD significantly outperform the baselines across datasets and metrics. 

As it can be seen, CVAE is a very strong baseline and outperform the other baselines in most situations. Compared with CDL, it can be seen that the inference network learns a better probabilistic representation of a latent variable for auxiliary information than CDL, leading to better performance. While CDL need additional noisy criteria in auxiliary information observation space, which makes it not robust. The inferior results of DeepCoNN may be due to that it only uses a single learner to learn user/item representations, only with auxiliary information as input compared to hybrid model. Therefore it cannot capture implicit relationships between users stored in interaction data very well. 

To focus more specifically on the comparison of CVAE and VCM, we can see that although both CVAE and VCM use deep learning models to extract representation for auxiliary information, the proposed VCM achieves better and more robust recommendation,
especially for large $R$. This is because VCM learns the user probabilistic representation by two stream VAEs set up, instead of learning the user/item latent factor through the point estimate of PMF. Besides, the collaborative learning mechanism allows the model to fully leverage the Bayesian deep representation from two views of information and lets the two learners be optimized synchronously. On the other hand, due to the point nature of the latent factor learned by PMF and alternative optimization, CVAE fails to achieve this robust performance. VCM-CD that uses the cross-domain inference to make the prediction can achieve better performance than VCM cause the review text we used here contains more specific information about users preference when the interaction data is extremely sparse. This promotion is especially obvious in the most sparse Clothing dataset. 
\subsection{Ablation Study}
In this subsection, we do the ablation study to understand better how the collaborative learning mechanism work, we develop:
\begin{itemize}
\item \textbf{VCM-Se}: The collaborative learning mechanism of VCM is removed. And the VCM is \textbf{se}parated as two independent variational models.
\item \textbf{VCM-OD}: We first train $\textrm{VAE}_{\textrm{y}}$ on the reviews alone without the influence of $\textrm{VAE}_{\textrm{x}}$. Then we fix $\textrm{VAE}_{\textrm{y}}$, and train $\textrm{VAE}_{\textrm{x}}$ with ${KL}(q_{\phi_{x}}(z_{u}|x_{u})||q_{\phi_{y}}(r_{u}|y_{u}))$. This means the information only can flow from $\textrm{VAE}_{\textrm{y}}$ to $\textrm{VAE}_{\textrm{x}}$ in \textbf{o}ne \textbf{d}irection which is different with the bi-directional flow in collaborative learning mechanism.
\item \textbf{VCM-NV}: The bi-directional KL regularization in collaborative learning mechanism is replaced with a constraint: $||\mu_{\phi_{x}}-\mu_{\phi_{y}}||_{2}^{2}$, which does \textbf{n}ot consider the \textbf{v}ariance $\sigma_{\phi_{x}}$ and $\sigma_{\phi_{y}}$ of the probabilistic representations.
\end{itemize}

The performance of VCM and its variants on Movies, Yelp, Clothing are given in Table~\ref{tab::ablation}. To demonstrate that the cooperation between two VAE can enhance recommendation performance, VCM-Se uses two independent VAEs for training without the collaborative learning mechanism. In this manner, we learn the two variational distribution $q_{\phi_{x}}$ and $q_{\phi_{y}}$ without considering the informative information from each other. As it can be seen in Table~\ref{tab::ablation}, VCM achieves the best performance. It verifies that modeling users' preference from two views does augment the performance of $\textrm{VAE}_{\textrm{x}}$. To investigate the importance of the bi-directional information flow in collaborative learning mechanism, VCM-OD is introduced that only consider one-directional information flow. Moreover, the performance gap between VCM-OD and VCM suggests that using collaborative synchronous training scheme is better than only using $\textrm{VAE}_{\textrm{y}}$ to enhance $\textrm{VAE}_{\textrm{x}}$. Furthermore, although VCM-NV can also learn probabilistic representation for two views of data, this constraint that is without considering variance makes two learners can not leverage all information stored in representations, leading the performance VCM-NV drops like CVAE with the same reason. 
\begin{table}[]
\centering
\caption{Comparing variants of the proposed model on the performance of NDCG@10. The best results are indicated in bold. $\dag$: p $<$ 0.01 in a statistical significance test, compared to the best variant.}
\begin{tabular}{|c|c|c|c|}
\hline
Model                  & Yelp             & Clothing         & Movies           \\ \hline
VCM-Se      & 0.036           & 0.015          & 0.046          \\ \hline
VCM-OD & 0.047          & 0.025           & 0.047          \\ \hline
VCM-NV            & 0.044          & 0.024          & 0.051          \\ \hline
VCM               & \textbf{0.051$\dag$}          & \textbf{0.027$\dag$}          & \textbf{0.057$\dag$} \\ \hline
\end{tabular}
\label{tab::ablation}
\end{table}
\subsection{The impact of collaborative learning on $\textrm{VAE}_\textrm{x}$}
It is natural to wonder how the collaborative learning promote the performance of $\textrm{VAE}_{\textrm{x}}$.
Intuitively, by modeling users latent variable with click behavior and review text collaboratively, VCM can learn the more expressive representation than VCM-Se. Therefore VCM could be more robust when user's click behavior data is scarce. To study this, we break down users into five groups based on their activity level. The activity level represents the number of items each user has clicked on. According to complementary slackness KKT condition~\cite{kuhn1951aw,karush1939minima}, we can use $KL(q_{\phi_{x}}(z_{u}|x_{u})||p_{z_{u}})$ as the approximation of the capacity limitation $\hat{c}_{u}$ after optimization. It indicates the amount of information stored in the variational distribution. We compute NDCG@$10$ and $\hat{c}_{u}$ for each group using VCM and VCM-Se. Table.\ref{tab::different_activity_level} summarizes how performance differs across users of different active levels.

It is interesting to find that, as the activity level increase, the variational distribution capacity of VCM and VCM-Se also monotonically increase. This phenomenon shows that, by using $\textrm{VAE}_{\textrm{x}}$ to learn the probabilistic representation of user latent variable, both VCM and VCM-Se can automatically allocate a proper user-level capacity for users of different levels to store the information. 

We can also find that the variational distribution capacity of VCM is all greater than VCM-Se for users of different levels in the three data sets. This shows the collaborative learning mechanism allows the information in review text flows from $q_{\phi_{y}}$ to $q_{\phi_{x}}$, which makes $q_{\phi_{x}}$ more expressive with more information, and then the $\textrm{VAE}_{\textrm{x}}$ automatically allocates more capacity to store the more comprehensive information. The promotion of capacity between the two models is particularly prominent for users who only click a small number of items~(shown in bold in Table~\ref{tab::different_activity_level}).
\begin{table}[]
\caption{NDCG@10 and approximation of Capacity $\hat{c}_{u}$ for users with increasing level of activity, and the activity level is measured by how many items a user clicked on. The larger the value of $\hat{c}_{u}$ is, the more information  the distribution $q_{\phi_{x}}$ contains. Although details vary across datasets, \textbf{VCM} consistently improves NDCG@10 and $\hat{c}_{u}$ for the user of different levels. The relative improvement is shown in bracket.}

\begin{tabular}{|c|c|c|c|c|}
\hline
                & \multicolumn{2}{c|}{\textbf{VCM-Se}} & \multicolumn{2}{c|}{\textbf{VCM}} \\ \hline
\textbf{level}  & \textbf{$\hat{c}_{u}$}           & \textbf{NDCG}          & \textbf{$\hat{c}_{u}$}      & \textbf{NDCG}      \\ \hline
\multicolumn{5}{|c|}{\textbf{Yelp}}                                                                        \\ \hline
\textbf{5-20}   & 9.7                    & 0.032                  & \textbf{18.3(+87\%)}       & \textbf{0.048(+49\%)}       \\ \hline
\textbf{20-40}  & 18.4                   & 0.051                  & 32.5(+76\%)       & 0.064(+26\%)       \\ \hline
\textbf{40-60}  & 25.2                   & 0.049                  & 42.9(+70\%)       & 0.064(+30\%)       \\ \hline
\textbf{60-80}  & 33.7                   & 0.049                  & 49.7(+47\%)       & 0.069(+42\%)       \\ \hline
\textbf{80-max} & 47.1                   & 0.093                  & 59.0(+25\%)       & 0.107(+15\%)       \\ \hline
\multicolumn{5}{|c|}{\textbf{Clothing}}                                                                    \\ \hline
\textbf{5-6}    & 6.7                    & 0.015                & \textbf{9.0(+33\%)}        & 0.027(+70\%)      \\ \hline
\textbf{6-7}    & 8.4                    & 0.016                & 10.7(+27\%)       & 0.031(+85\%)       \\ \hline
\textbf{8-9}    & 10.3                   & 0.017                & 12.7(+23\%)       & \textbf{0.034(+94\%)}       \\ \hline
\textbf{10-11}  & 11.0                   & 0.022                & 13.4(+21\%)       & 0.036(+55\%)       \\ \hline
\textbf{12-max} & 14.4                   & 0.023                & 17.4(+20\%)       & 0.036(+53\%)       \\ \hline
\multicolumn{5}{|c|}{\textbf{Movies}}                                                                      \\ \hline
\textbf{5-20}  & 10.0                   & 0.046                  & \textbf{19.7(+96\%)}       & 0.056(+22\%)       \\ \hline
\textbf{20-40}  & 20.9                   & 0.046                  & 33.4(+59\%)       & 0.059(+27\%)       \\ \hline
\textbf{40-60}  & 30.1                   & 0.048                  & 43.8(+45\%)       & \textbf{0.062(+30\%)}       \\ \hline
\textbf{60-80}  & 36.3                   & 0.051                  & 50.4(+38\%)       & 0.064(+25\%)       \\ \hline
\textbf{80-max} & 66.2                   & 0.087                  & 72.1(+8\%)        & 0.100(+14\%)       \\ \hline
\end{tabular}
\label{tab::different_activity_level}
\end{table}
\subsection{The impact of collaborative learning on $\textrm{VAE}_\textrm{y}$}
The multinomial distribution $p_{u}$ is to model the probability of each word appearing in the review document $y_{u}$ for user $u$. Without collaborative learning, the likelihood of the review document rewards the $\textrm{VAE}_\textrm{y}$ for only putting probability mass on the high-frequency words in $y_{u}$. However, with the influence of $\textrm{VAE}_\textrm{x}$ under the collaborative learning mechanism, $\textrm{VAE}_\textrm{y}$ should also assign more probability mass on the keywords that can represent user preference.

\begin{figure}
    \centering
    \includegraphics[width=.45\textwidth]{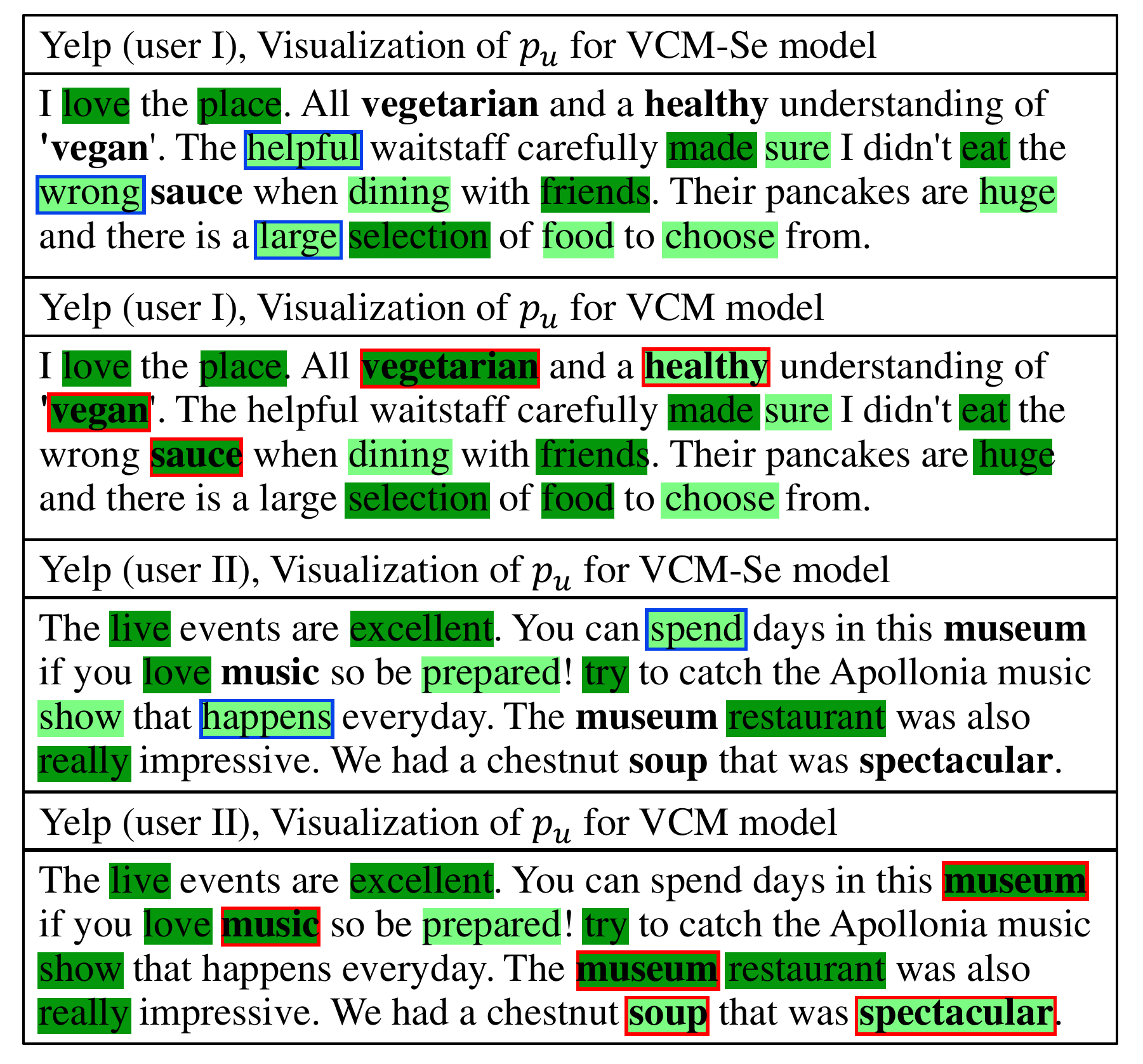}
    \caption{Highlighted words by $p_{u}$ in two users' review}
    \label{fig:visulization_topic_model}
\end{figure}

We highlight words that have high probability $p_{uv}$ in Figure~\ref{fig:visulization_topic_model}. We randomly sample the review example from two users in Yelp dataset. Words with the highest probability are colored dark-green, high probability words are lighted-green, and low/medium probability words are not colored. In Figure~\ref{fig:visulization_topic_model}, we compare the $p_{u}$ of VCM-Se and VCM model. For convenient comparison, we use blue and red rectangles to emphasize their differences.

For User I, $\textrm{VAE}_\textrm{y}$ of VCM puts more probability on "vegetarian," "healthy," "vegan," "sauce" words which show that the user may be a vegetarian and put more attention on healthy habit. While, without the collaborative learning mechanism, $\textrm{VAE}_\textrm{y}$ of VCM-Se puts more probability on some meaningless words such as "helpful," "wrong," "large." A similar result is observed for user II. The words "music" and "museum" show the obvious preference. This demonstrates the collaborative learning mechanism has a beneficial influence on both two learners, which not only can enhance the recommendation performance for $\textrm{VAE}_\textrm{y}$ but also make $\textrm{VAE}_\textrm{y}$ capture the more representative words.
\section{Related work}
Compared to the CF-based approach, the hybrid model relies on only two sources of information to mitigate the sparsity problem. Based on the how tightly the interaction data and auxiliary information are integrated, the hybrid model can be divided into two subcategories: loose coupled and tightly coupled methods~\cite{Wang:2015:CDL:2783258.2783273}. The loosely coupled method combines the output from separate collaborative and content-based systems into a final recommendation by a linear combination~\cite{miranda1999combining} or voting scheme~\cite{pazzani1999framework}. The tightly coupled method takes the processed auxiliary information as a feature of the collaborative method~\cite{li2011generalized}. However, they all assume that the features are the good representation which is usually not the case. Collaborative topic regression~(CTR)~\cite{Wang:2011:CTR:2020408.2020480} is a method that explicitly integrates the latent Dirichlet allocation~\cite{blei2003latent}~(LDA) and PMF for two source information with promising result. However, the representation ability is limited to the topic model. 

On the other hand, deep learning model has shown great potential for learning effective representations~\cite{vincent2010stacked}. Very recently, Neural collaborative filtering~\cite{He:2017:NCF:3038912.3052569} and VAE-CF~\cite{Liang:2018:VAC:3178876.3186150} that use neural networks have shown the promising result, but they belong to CF-based methods. CDL~\cite{Wang:2015:CDL:2783258.2783273} and collaborative recurrent autoencoder have been proposed for joint learning a SDAE~\cite{vincent2010stacked}~(a denoising recurrent autoencoder) with PMF. Both of the models try to learn representation from auxiliary information with additional denoising criteria. To make a further step, CVAE propose to infer the stochastic distribution of the latent variable through the neural network for auxiliary information. However, most previous works use an asynchronous mutual regularization between learners which cannot fully leverage the representations for two sources of information.

There is also another line of research that only utilizes one single learner with only auxiliary information such as review text as input for rating regression~\cite{Chen:2018:NAR:3178876.3186070,seo2017interpretable}, DeepCoNN~\cite{Zheng:2017:JDM:3018661.3018665} that models users and items using review text for rating prediction problems have shown promising result. Although they utilize word-embedding technique~\cite{Mikolov:2013:DRW:2999792.2999959} and Convolutional neural network~\cite{collobert2011natural}~(CNN) to learn good representation for text data, compared to the hybrid model, it only uses one single learner to learn user/item representation only with the auxiliary information as input, so it can not capture the implicit relationship between users stored in interaction data well.

\section{Conclusion}
This paper proposes the variational collaborative model that jointly model the generation of auxiliary information and interaction data collaboratively. It is a deep generative probabilistic model that learns a probabilistic representation of user latent variable through VAE, leading to robust recommendation performance. To the best of our knowledge, VCM is the first pure deep learning model that can fully leverage the probabilistic representation learned from different sources of data due to the synchronous collaborative learning mechanism. The experiment has shown that the proposed VCM can significantly outperform the state-of-the-art methods for the hybrid recommendation with more robust performance.

\bibliography{formatting-instructions-latex-2019}
\bibliographystyle{aaai}

\end{document}